\documentclass[conference]{IEEEtran}
\IEEEoverridecommandlockouts
\usepackage{cite}
\usepackage{amsmath,amssymb,amsfonts,amsthm}
\usepackage[ruled,vlined]{algorithm2e}
\usepackage{algorithmic}
\usepackage{graphicx}
\usepackage{textcomp}
\usepackage{xcolor}
\def\BibTeX{{\rm B\kern-.05em{\sc i\kern-.025em b}\kern-.08em
    T\kern-.1667em\lower.7ex\hbox{E}\kern-.125emX}}

\theoremstyle{definition}
\newtheorem{definition}{Definition}
\newtheorem*{problem}{Problem Definition}

\definecolor{myyellow}{RGB}{242,176,61}
\definecolor{myblue}{RGB}{62,147,242}

\definecolor{burgundy}{rgb}{0.5, 0.0, 0.13}

\begin{document}

\title{Pay Attention to Relations: Multi-embeddings for Attributed Multiplex Networks}

\author{\IEEEauthorblockN{Joshua Melton\IEEEauthorrefmark{1},
Michael Ridenhour\IEEEauthorrefmark{2}, and 
Siddharth Krishnan\IEEEauthorrefmark{3}}
\IEEEauthorblockA{\textit{Dept. of Computer Science} \\
\textit{University of North Carolina at Charlotte}\\
Charlotte, USA \\
\IEEEauthorrefmark{1}jmelto30@uncc.edu,
\IEEEauthorrefmark{2}mridenh7@uncc.edu,
\IEEEauthorrefmark{3}skrishnan@uncc.edu}
}

\maketitle

\begin{abstract}
Graph Convolutional Neural Networks (GCNs) have become effective machine learning algorithms for many downstream network mining tasks such as node classification, link prediction, and community detection. However, most GCN methods have been developed for homogenous networks and are limited to a single embedding for each node. Complex systems, often represented by heterogeneous, multiplex networks present a more difficult challenge for GCN models and require that such techniques capture the diverse contexts and assorted interactions that occur between nodes. In this work, we propose \textbf{RAHMeN}, a novel unified {\bf r}elation-aware embedding framework for {\bf a}ttributed {\bf h}eterogeneous {\bf m}ultipl{\bf e}x {\bf n}etworks. Our model incorporates node attributes, motif-based features, relation-based GCN approaches, and relational self-attention to learn embeddings of nodes with respect to the various relations in a heterogeneous, multiplex network. In contrast to prior work, RAHMeN is a more expressive embedding framework that embraces the multi-faceted nature of nodes in such networks, producing a set of multi-embeddings that capture the varied and diverse contexts of nodes.
 
We evaluate our model on four real-world datasets from Amazon, Twitter, YouTube, and Tissue PPIs in both transductive and inductive settings. Our results show that RAHMeN consistently outperforms comparable state-of-the-art network embedding models, and an analysis of RAHMeN's relational self-attention demonstrates that our model discovers interpretable connections between relations present in heterogeneous, multiplex networks.
\end{abstract}


\section{Introduction}
Deep graph neural network (GNN) embeddings have become increasingly successful in graph-based learning tasks like node classification, link prediction, and network reconstruction~\cite{graphsage,kipf_gcn,deepwalk,node2vec,cui2020adaptive,gat}. While graph convolutional networks (GCNs) have been instrumental in the development of the embedding techniques, most proposed works in the GCN family consider only \emph{homogeneous} networks, which consist of a single node and edge type. However, real-world networks are often complex, and their properties are characterized via multiple attributed edge and node types---examples include gene-protein interaction networks and knowledge graphs. Nodes in such networks participate in a variety of different contexts and relate to one another in different ways. This multi-faceted nature of nodes in such networks belies the limited expressiveness of GCN models designed for homogeneous networks. Therefore, to accomplish downstream learning tasks on such attributed, heterogeneous, multiplex networks, we require network embedding techniques with a greater expressive capacity. Such frameworks must capture the varied contextual features along with differing local structural topology of nodes in these networks in order to best perform predictive and classification tasks in downstream graph learning objectives.

\begin{figure}[!ht]
  \centering
  \includegraphics[width=8cm]{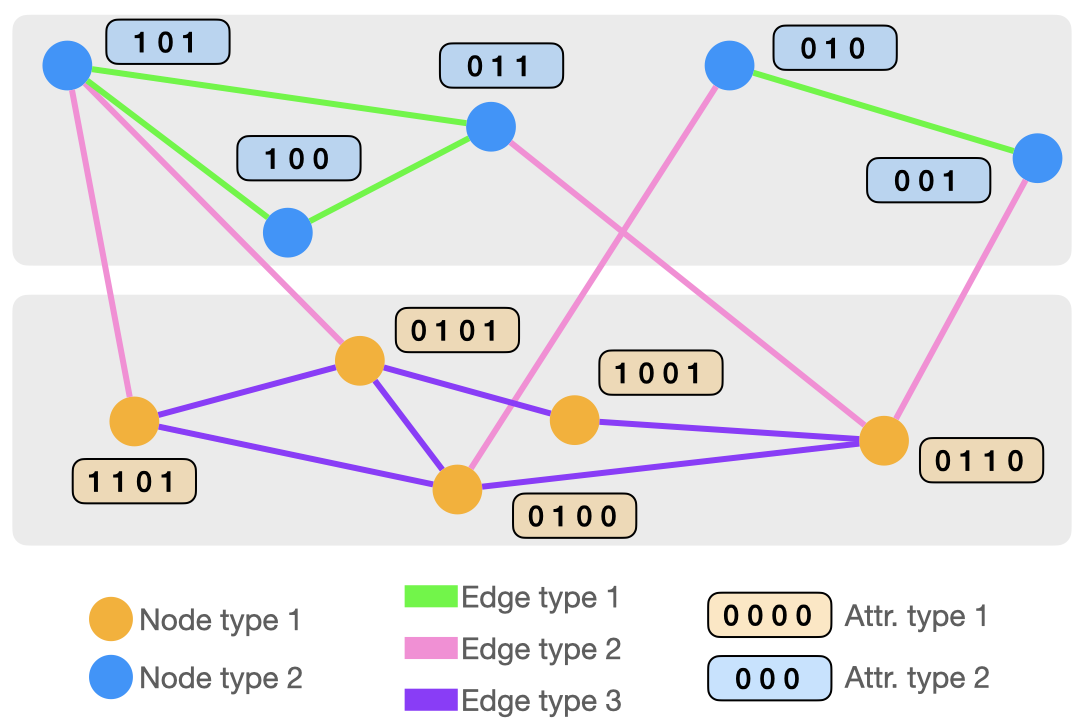}
   \caption{\small Example \textbf{Attributed Heterogeneous Multiplex Network} with multiple types of nodes and edges. Each node type consists of their own set of attributes — \textcolor{myblue} {\textbf{blue}} nodes with 3 attributes and \textcolor{myyellow} {\textbf{yellow}} nodes with 4 attributes. }
  \label{fig:example_hetnet}
\end{figure}

Towards that end, in this work we propose {\em a unified network embedding framework for real-world complex structures---networks that are attributed, multiplex, and heterogeneous.} Recent research studies focus on developing network embedding models for multiplex networks (single node type and multiple edge types)~\cite{Zitnik2017,mne,multinet}, heterogeneous networks (multiple node and edge types)~\cite{metapath2vec,shi2018heterogeneous}, and heterogeneous multiplex networks (multi-level network with multiple node and edge types)~\cite{rgcn,gatne,han}. Our present work introduces a more expressive embedding framework by asserting the \emph{multi-faceted nature} of nodes throughout the embedding process and eschewing the reliance on a single embedding as insufficient for capturing a node's proximity to others in such complex networks. We therefore design a novel graph convolutional approach for network embedding with \emph{relational self-attention} that produces a set of information-rich \emph{multi-embeddings} of nodes present in an Attributed Heterogeneous Multiplex Network (AHMeN) (example shown in Figure~\ref{fig:example_hetnet}). An instance of an AHMeN appears, for example, in protein-protein interaction networks~\cite{halu2019multiplex} where a set of proteins interact in a number of different tissues. By incorporating the interactions and protein features across multiple tissue layers in the network, our embedding framework can better capture the many-sided functional roles of a protein and can better predict unseen protein-protein interactions in various tissue layers.

\textbf{Present work}: In this work, we present RAHMeN---Relation-aware Embeddings for Attributed Heterogeneous Multiplex Networks---an expressive and interpretable framework to learning spatial embeddings of nodes in AHMeNs. We make the following contributions and observations to the current line of research on network embedding with RAHMeN:
\begin{enumerate}
	\item We employ a set relational graph convolutional operators that incorporate a relation-specific view of a node's self-information with the node's local relation-specific neighborhood, providing an enhanced context of the node in its neighborhood with respect to each relation in the network.
	\item To the sequence of relation-specific node representations, we apply semantic self-attention to share information across each relation type in the network, informing the node's context across all relation-types present in the network. This produces a set of robust multi-embeddings for the node characterizing its various contexts in the network.
	\item The latent representations learned with our framework give state-of-the-art performance on both transductive and inductive link prediction tasks with four real-world datasets.
	\item We propose to incorporate higher-order subgraph features as node attributes. We observe that such structural features can enhance the node representations learned by the framework and allow for inductive learning on networks without node attributes.
	\item Our self-attention mechanism uncovers interpretable connections between relations in the network. For example, we show that our model confirms domain knowledge by establishing connections between proteins found in the brain, central nervous system, and nervous system.
\end{enumerate}

\section{Related Work}
 
\noindent{\textbf{Network Embeddings.}} Traditional network embedding models~\cite{node2vec,deepwalk} use random walks to preserve the local and global neighborhoods of a node. Advances in deep graph neural networks have introduced the message-passing paradigm~\cite{gilmer_msgpass} and convolutional graph embedding models, like Graph Convolutional Networks (GCN)~\cite{kipf_gcn} and graphSAGE~\cite{graphsage}, utilize node attributes and neighborhood aggregation to capture a node's local context. Graph Attention Networks (GAT)~\cite{gat} further improved such graph convolutional approaches by introducing node-level attention mechanisms during neighbor aggregation. These frameworks wed spatial and spectral interpretations of networks by incorporating node attributes and the contextual information of a node's local receptive field, defined by the node's neighborhood in the graph~\cite{kipf_gcn,cui2020adaptive}, with learnable filter functions providing efficient implementations and generating node embeddings with provable representation power~\cite{kipf_gcn,graphsage,multisage}. Parameter sharing allows such models to scale to large datasets and to be applied in inductive contexts~\cite{graphsage,ying2018graph}.

\noindent{\textbf{Heterogeneous and Multiplex Graph Embeddings.}} Such convolutional approaches were originally limited to homogeneous networks, but in recent years a number of heterogeneous, multiplex network embedding frameworks have been proposed to extend graph neural network embedding models to such networks. Frameworks like R-GCN~\cite{rgcn} and HAN~\cite{han} extend the idea of GCN~\cite{kipf_gcn} and GAT~\cite{gat}, respectively, for heterogeneous networks by applying relation or metapath-specific convolutional operations. Frameworks like MNE~\cite{mne} and GATNE~\cite{gatne}, employ a base node embedding which is augmented by edge-type specific messages generated by graph convolutional operations. HAN and GATNE have both employed semantic attention~\cite{selfattn} as a means to implicitly learn aggregation weights for edge-type or metapath specific embeddings, respectively.

To date, most frameworks consider only a single embedding for each node~\cite{mne,rgcn,han} and do not sufficiently capture the variety of contexts and interactions between nodes in heterogeneous, multiplex networks. In RAHMeN, we emphasize that nodes in such real-world networks are related to one another in different ways and thus cannot be accurately modeled by a single embedding. We therefore propose a more expressive multi-embedding framework for nodes in heterogeneous, multiplex networks called RAHMeN that incorporates a set of relation-specific graph convolutional operators, which learn the individual semantic contexts for each node, and a relational self-attention mechanism, which shares information across a node's various relational contexts, to produce a set of robust multi-embeddings for each node in a heterogeneous, multiplex network.

\section{Problem Definition}

The complete set of notations used in this paper is given in Table \ref{tab:notations}. We define a homogeneous network as $G = (\mathcal{V}, \mathcal{E})$, where $\mathcal{V}$ denotes the set of all nodes in the graph, and $\mathcal{E}$ denotes the set of all edges in the graph. Based on this fundamental representation of a homogeneous network, we define the following:

\begin{definition}[Attributed Network]
  We define an attributed homogeneous network as $G = (\mathcal{V}, \mathcal{E}, \mathcal{A}) $, where each node $ v \in \mathcal{V} $ is associated with a set of node features $ \mathcal{A} = \{\boldsymbol{x_i} \, | \, v_i \in \mathcal{V} \} $, where $ \boldsymbol{x_i} $ is the feature vector associated with node $ v_i $.
\end{definition}

\begin{definition}[Heterogeneous \& Multiplex Network]
  A heterogeneous network is defined as $ G = (\mathcal{V}, \mathcal{E}, T_v, T_e) $, where $ \mathcal{V}$ and $ \mathcal{E}$ are the universal sets of nodes and edges. Each node $ v \in \mathcal{V}$ is associated with mapping function $ \phi(v): \mathcal{V} \to T_v $, and each edge $ e \in \mathcal{E} $ is associated with mapping function $ \psi(e): \mathcal{E} \to T_e $, where $ T_v $ and $ T_e $ denote the sets of node types and edge types, respectively. If $ |T_v| + |T_e| > 2 $, the network is termed \textbf{heterogeneous}. The network is termed \textbf{multiplex} if multiple types of edges may exist between a pair of nodes.
\end{definition}

\begin{figure}
	\centering
	\includegraphics[width=6cm]{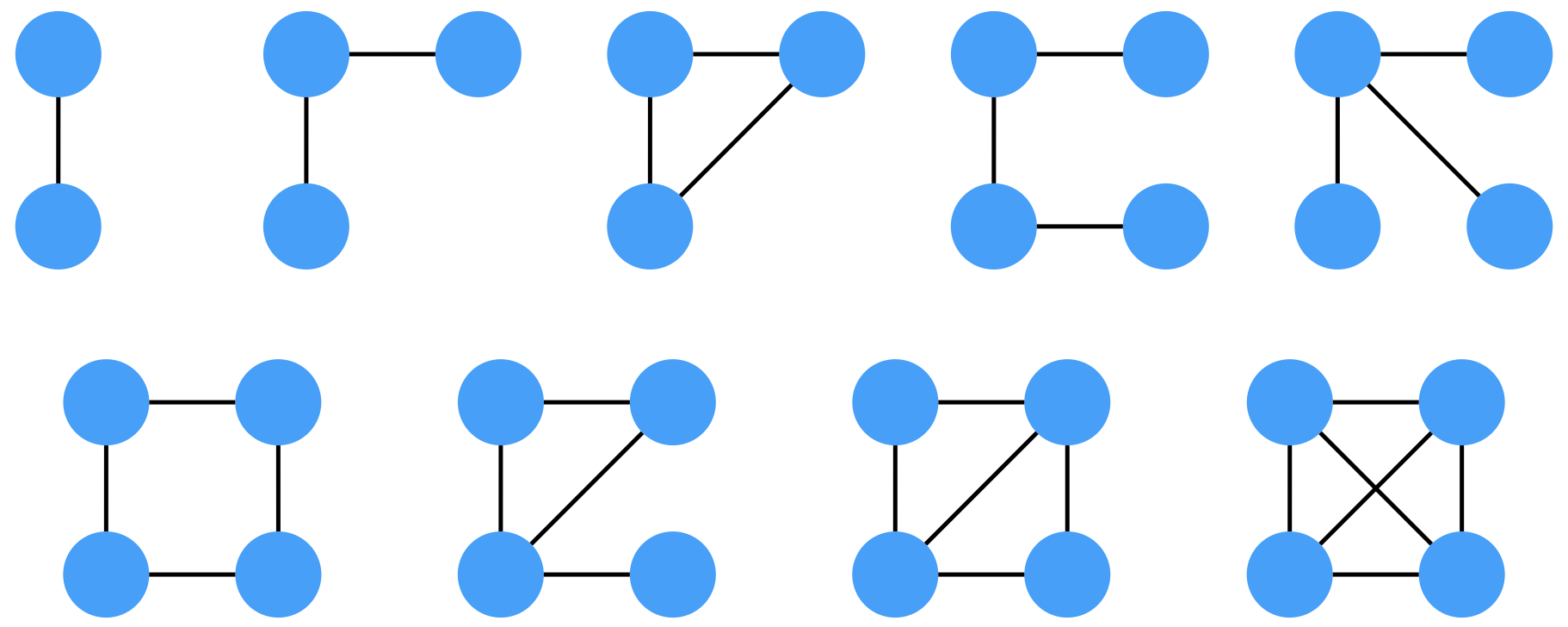}
	\caption{\small Two --- four node connected motifs considered for structural feature representations.}
	\label{fig:motifs}
\end{figure}

\begin{definition}[Motif]
  Given a graph $G = (\mathcal{V}, \mathcal{E})$, a motif is defined as a subgraph $M = (\mathcal{V}_m, \, \mathcal{E}_{m})$ consisting of $m$ nodes $\mathcal{V}_m \subset \mathcal{V}$ and with all edges $\mathcal{E}_m \subset \mathcal{E}$ connecting $\mathcal{V}_m$ nodes in $G$ \cite{hone}. For this work, we consider two--four node connected motifs shown in Figure \ref{fig:motifs}.
\end{definition}

\begin{definition}[Canonical Relation]
  Given a heterogeneous graph $ G $ with $ T_v $ node types and $ T_e $ edge types, we define the canonical relation $r$ as a 3-tuple representing the source node type, edge type, and destination node type of the relation. The set $R$ consists of all canonical relations $ \{r = (t_{v}^i, \: t_{e}^{(i,j)}, \: t_{v}^j) \: | \:  \{t_v^i, \: t_v^j\} \in T_v, \: {T_e^{(i,j)} \in T_e\}} $ and uniquely defines the set of relations present in the network.
\end{definition}

\begin{problem}[AHMeN Embedding]
  Given an AHMeN, $ G = (\mathcal{V}, \mathcal{E}, \mathcal{A}, R) $, where $\mathcal{V}$ is a set of vertices, $\mathcal{E}$ is a set of edges, $\mathcal{A}$ is a set of node attributes, and $R$ is a set of all canonical relations, the goal is to learn a transformation function that gives a set of low-dimensional representations for each node $v \in \mathcal{V}$ with respect to canonical relations $ r \in R $, such that nodes which are similar are closer to one another in the embedding space. That is, learn a function

  \begin{center}
    $ f(v): \mathcal{V} \to \mathbb{R}^{d} \; \forall \, r \in R $, where $ d \ll |\mathcal{V}| $
  \end{center}
\end{problem}

\begin{table}
  \centering
  \caption{Important notations and their definitions}
  \begin{tabular}{l|l}
    \textbf{Notation} & \textbf{Description} \\
    \hline
    $ G $ & the input network \\
    $ \mathcal{V}, \mathcal{E} $ & the node and edge sets of $ G $ \\
    $ T_v, T_e $ & the node-type and edge-type sets of $ G $ \\
    $ \mathcal{A} $ & the attribute set of $ G $ \\
    $ R $ & the set of canonical relations of $ G $ \\
    $ v, e $ & a node and edge in the graph \\
    $ \boldsymbol{x} $ & the set of attributes of a node \\
    $ \mathcal{N} $ & the neighborhood of a given node \\
    $ k $ & neighborhood levels (or) hops \\
    $ z $ & the latent representation of a node \\
    $ d $ & the dimension of the final overall embeddings \\
    $ d_a $ & the dimension of the attention vector \\
    $ \boldsymbol{\textrm{W}_{s}}, \boldsymbol{\textrm{W}_{n}} $ & the self and neighbor transformation matrices \\
  \end{tabular}
  \label{tab:notations}
\end{table}

\begin{figure*}[!ht]
  \centering
  \includegraphics[scale=0.25]{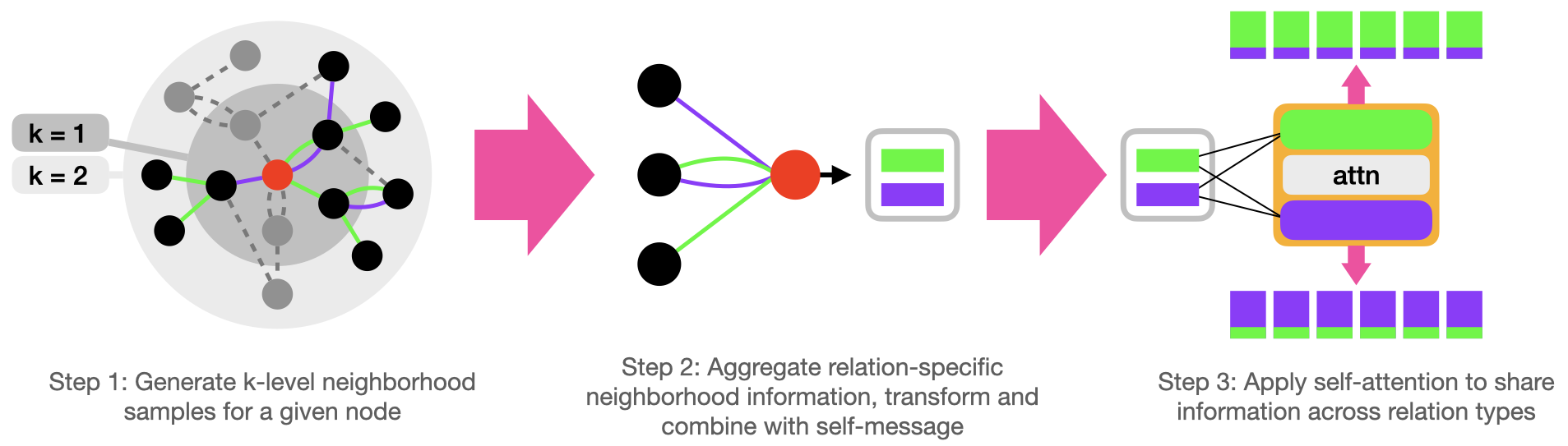}
  \caption{Overview of steps proposed in RAHMeN. In step 1, K-level neighborhood samples are taken for the target node. For each k-level neighborhood sample: In step 2, relation-specific graph convolutions incorporate a relational view of the node's self-attributes with its local relation-specific neighborhood. In step 3, self-attention is applied to the sequence of latent representations to share information across relations in the network. Steps 2 and 3 are repeated for each neighborhood level up to K to produce a set of embeddings $z$ for the target node.}
  \label{fig:pipeline}
\end{figure*}

\section{Methodology}
In this section, we describe the proposed RAHMeN framework. We explain the model architecture, its ability to be applied in an inductive context, and describe an efficient semi-supervised method for model optimization. The model pipeline is illustrated in Figure \ref{fig:pipeline}.

\subsection{RAHMeN Framework}

In the RAHMeN framework, we learn an inductive transformation function for each node $v$ with respect to all relations in the network, defined by the set of canonical relations $R$. This transformation function produces a set of low-dimensional spatial representations of a node by jointly incorporating relation-specific views of the self-node's attributes with its multiple network contexts, defined by canonical relations $R$. At each k-level--corresponding to the k-hop neighborhood of node $v$---we apply relational graph convolutional operations over the views of the network $ G_r = (\mathcal{V}_r, \mathcal{E}_r, \mathcal{A}) | \forall r \in R $. We then apply a relational semantic attention mechanism to differentiate and optimally combine the semantic-specific latent representations of node $v$.

The initial representation $ \boldsymbol{h}_v^0 $ for each node $ v $ is given by the node features vector $ \boldsymbol{x}_v $. Following the message-passing paradigm described in \cite{gilmer_msgpass}, our neighbor message function is defined as:

\begin{equation}
    \phi_r^k \left( \mathcal{N}(v,r) \right) = \sum_{u \in \mathcal{N}(v,r)} \frac{1}{|\mathcal{N}(v,r) |} \, \boldsymbol{\mathrm{W}}_{n, r}^k \, \boldsymbol{h}_{u, r}^{k-1}
\label{eq:msgfn}
\end{equation}

\noindent
where $\mathcal{N}(v,r)$ is node $v$'s local neighborhood on relation $r$, and $\boldsymbol{h}_{u,r}^{k-1}$ is each node $u$'s latent representation from the previous layer.

The transformed and aggregated neighborhood message is then combined with a specific view of node $v$'s self-representation to generate a relation-specific latent representation of node $v$:

\begin{equation}
\boldsymbol{\tilde{h}}_{v,r}^k = \sigma \left( \boldsymbol{\mathrm{W}}_{s, r}^k \, \boldsymbol{h}_{v, r}^{k-1} + \phi_r^k \left( \mathcal{N}(v,r) \right) + \boldsymbol{b}_r^k \right)
\label{eq:conv}
\end{equation}

\noindent
where $\sigma$ is a non-linear activation function, such as ELU~\cite{elu}. The self and neighbor transformation functions, $\boldsymbol{\mathrm{W}_s}$ and $\boldsymbol{\mathrm{W}_n}$ may be any differentiable function such as a linear transformation or an MLP.

Every node in an AHMeN participates in multiple semantic relationships defined across the various relations in the network. The latent representations of node $v$ along each relation $r \in R$ reflect only one aspect of the node's semantic context and do not take into account interactions between the various relations in the network. To learn a more comprehensive set of embeddings for a node, we apply semantic attention~\cite{selfattn} to this sequence of latent node representations. An emphasis of RAHMeN is the irreducibility of a node's complex position in an AHMeN to a single embedding. As such, we employ our relational semantic self-attention to generate a set of multi-embeddings for a node, which blend the various relational contexts of a node while preserving a focus on a particular relation within each individual embedding.

First, we stack each of the relation-specific representations of node $v$ from equation~\ref{eq:conv} as the sequence below, with shape $ |R| \times d $:

\begin{equation}
    \boldsymbol{\tilde{h}}_{v}^{k} = \textsc{concat}(\boldsymbol{\tilde{h}}_{v,r}^k \: | \: \forall \, r \in R)
\end{equation}

To learn the optimal set of attention weights for each relation with respect to every other relation in the network, we transform the embeddings using a nonlinear transformation and compute the similarity with a relation-level attention matrix. The attention weights for each relation are obtained by normalizing the above results using the softmax function.

\begin{equation}
    \boldsymbol{a}_{v}^k = \textrm{softmax}(\boldsymbol{\mathrm{W}}_{rel}^k \cdot \textrm{tanh}(\boldsymbol{W}_{attn}^k \cdot \boldsymbol{\tilde{h}}_v^k))
\end{equation}

\noindent
where $ \boldsymbol{\mathrm{W}}_{rel}^k $ is a trainable relation attention matrix of shape $ |R| \times d_a \times |R| $ and $ \boldsymbol{W}_{attn}^k $ is a trainable transformation matrix with size $ |R| \times d \times d_a  $. The final latent representation for node $v$ at level $k$ is therefore:

\begin{equation}
    \boldsymbol{h}_v^k = \boldsymbol{a}_v^k \cdot \left[ \boldsymbol{\tilde{h}}_{v,r}^k \,| \, \forall \, r \in R \right]
\label{eq:rahmen}
\end{equation}

\noindent
The overall set of multi-embeddings of node $v$ at $K$ with shape $ |R| \times d $ is
\begin{equation}
    \boldsymbol{z}_v = \boldsymbol{h}_v^K
\end{equation}

\subsection{Model Optimization} \label{modelopt}

In this section, we describe the semi-supervised training process for the proposed RAHMeN framework. Following~\cite{node2vec,gatne}, we use random walks to generate sequences of node sequences and optimize our model to learn node representations that maximize the similarity of co-occuring nodes. We conduct random walks along each relation view $G_r = (\mathcal{V}_r, \mathcal{E}_r, \mathcal{A}) $. Given schema $ \mathcal{M} : \mathcal{V}_1 \to \mathcal{V}_2 ... \mathcal{V}_m, ..., \to \mathcal{V}_l$, where $l$ is the length of the metapath schema. The transition probability at step $t$ is:

\begin{equation}
\centering
p(u \, | \, v, \mathcal{M}) = \begin{cases}
\frac{1}{| N(v, r) \, \cap \, \mathcal{V}_{t+1} |} & (v, u) \in r \\
\hspace{0.9cm} 0  & (v, u) \notin r\\
\end{cases}
\end{equation} 

where $ \mathcal{N}(v,r) $ denotes neighborhood of node $v$ along relation $r$. The random walker conducts a set of walks for each node along each relation $r$ in which it participates, capturing the various semantic contexts for each relation in the network. A random walk with length $l$ along relation $r$ defines a path $P = (v_{p_1}, v_{p_2}, ..., v_{p_l})$. Given a context sliding window size $c$, we define the context of $v$ as $C = \{v_{p_j} | v_{p_i} \in P, |i-j| \leq c, j \neq i \}$. Therefore, given a node $v$ with context $C$ along a path, the objective is to minimize the negative log-likelihood:

\begin{equation}
-log \, P_{\theta}(\{u \, | \, u \in C  \} \,| \, v) = \sum_{u \in S}  - log\: P_{\theta}(u \, | \, v)
\end{equation}

where $\theta$ represents the model parameters. We utilize the softmax function as the probability of node $u$ given $v$:

\begin{equation}
P_{\theta}(u | v) = \frac{ \textrm{exp}(\boldsymbol{c}_r^T \cdot \boldsymbol{z}_{v,r}) }{ \sum_{u \in V_c} \textrm{exp} (\boldsymbol{c}_u^T \cdot \boldsymbol{z}_{v,r}) }
\end{equation}

where $u \in V_m$, $\boldsymbol{c}_u$ is the context embedding of node $u$, and $z_{v,r}$ is the embedding for node $v$ on relation $r$. To approximate the objective function, we use negative sampling for each node context triple $(v, r, u)$ as:

\begin{equation}
E = - \textrm{log} \: \sigma ( c_u^T \cdot \boldsymbol{z}_{v,r} ) - \sum_{l=1}^{L} \mathbb{E}_{v_i \sim P_{r(u)}} [ \textrm{log} \: \sigma ( -c_u^T \cdot \boldsymbol{z}_{v,r} ) ]
\label{eq:loss}
\end{equation}

where $\sigma$ is the sigmoid function. $L$ corresponds to the number of negative samples draw for each positive sample and $v_i$ is a node drawn randomly from noise distribution $P_r(v)$ defined on node $u$'s corresponding node set, and $P_r(v)$ may be either a uniform distribution or a log-uniform distribution ordered by node degree. The time complexity of the random walk based training algorithm is $O( |\mathcal{V}|\cdot|R|\cdot d \cdot L )$ where $|\mathcal{V}|$ is the number of nodes, $|R|$ is the number of relations, $d$ is the embedding dimension, and $L$ is the number of negative samples per training sample. The memory complexity of our algorithm is $O(|\mathcal{V}| \cdot |R| \cdot d \cdot d_a)$.

% \begin{algorithm}[tb]
%     \caption{RAHMeN}
%     \label{alg:train}
%     \textbf{Input}: network $ G = (\mathcal{V}, \mathcal{E}, \mathcal{A}, R) $, embedding dimension $d$, learning rate $\eta$, negative samples $L$ \\
%     \textbf{Output}: overall embeddings $\boldsymbol{z}_{v}$ for all nodes on every relation type $r$
%     \begin{algorithmic}[1] %[1] enables line numbers
%         \STATE Initialize model parameters $\theta$
%         \STATE Generate random walks on each relation $r$ as $\mathcal{P}_r$
%         \STATE Generate training samples $(\{v, r, u\})$ from random walks $\mathcal{P}_r$ on each relation $r$.
%         \WHILE{not converged}
%         \FOR{ \textbf{each} $(v, r, u) \in$ training samples }
%         \STATE Calculate $\boldsymbol{z}_{v,r} $ using Equation \ref{eq:rahmen}
%         \STATE Sample $L$ negative samples and calculate objective function $\mathcal{O}$ using Equation \ref{eq:loss}
%         \STATE Update model parameters $ \theta $ by $ \frac{\partial \mathcal{O}}{\partial \theta} $
%         \ENDFOR
%         \ENDWHILE
%     \end{algorithmic}
% \end{algorithm}

\section{Experiments}

We evaluate our model in both transductive and inductive experiments against state-of-the-art network embedding methods. The code for our analysis is available here\footnote{https://anonymous.4open.science/r/rahmen-anon-B817/README.md}. Sections \ref{datasets} and \ref{baselines} outline the datasets and baseline models used for our analysis. Our experiments seek to answer the following questions:
\begin{enumerate}
	\item Can we utilize the RAHMeN framework for link prediction on four real-world datasets in a \textbf{transductive} setting? (Section \ref{transductive})
	\item Can we apply our model for link prediction in an \textbf{inductive} setting? (Section \ref{inductive})
	\item Does RAHMeN's self-attention discover explainable connections between relations in networks? (Section \ref{attention})
	\item Can we improve the performance of GNNs by using network motifs to enhance node features? (Section \ref{motifs})
	\item How sensitive is our model to various hyperparameters, including embedding dimension, neighborhood sample size, number of k-levels, and attention dimension? (Section \ref{params})
\end{enumerate}

\subsection{Datasets} \label{datasets}

We utilize four publicly available datasets for our experiments. Table \ref{tab:datasets} provides the network properties of our datasets. Descriptions of each dataset are as follows:

\textbf{Amazon}\footnote{http://jmcauley.ucsd.edu/data/amazon/} We utilize the dataset\footnote{https://github.com/THUDM/GATNE \label{gatne_repo}} as provided by \cite{gatne}, which consists of only the product metadata of the \textit{Electronics} category from Amazon.com. Each product has a set of attributes including price, sales-rank, brand, and category.

\textbf{Twitter}\footnote{https://snap.stanford.edu/data/higgs-twitter.html} The Twitter dataset contains tweets and user interactions related to the discovery of the Higgs boson in 2012. For this work, we extract the multiplex network consisting of the reply, retweet, and mention networks between all users that had at least one reply link in the original data.

\textbf{YouTube}\footnote{http://socialcomputing.asu.edu/datasets/YouTube} We utilize the YouTube dataset provided by \cite{gatne}, which consists of a multiplex network describing the co-occurence of friends, subscriptions, favorited videos, subscribers, and a layer representing contacts between users.

\textbf{Tissue-PPI}\footnote{http://snap.stanford.edu/ohmnet/} The tissue-specific protein interaction network consists of protein interactions for 107 tissue types (network layers) in the human body. We extract multiplex network consisting of the ten largest network layers present in the dataset.

\begin{table}
	\centering
	\caption{Properties of Heterogeneous Multiplex Network datasets used for evaluation of RAHMeN framework.}
	\begin{tabular}{l | rrrr}
		\textbf{Dataset} & \# nodes & \# edges & \# relations \\
		\hline
		\rule{0pt}{2ex} Amazon		& 10,099	& 135,761	& 2  \\
						Twitter		& 28,473	& 91,726	& 3	 \\
						YouTube		& 2,000		& 1,310,544	& 5	 \\
						Tissue-PPI	& 4,360		& 527,850	& 10 \\
	\end{tabular}
	\label{tab:datasets}
\end{table}

\begin{table*}[t]
\centering
\caption{Performance (ROC-AUC \% and F1 \%) of embedding frameworks on the link prediction task in a transductive context. RAHMeN outperforms all baseline models on the Amazon, YouTube, and Tissue PPI datasets (unpaired t-test, $p < 0.05$) and equals the performance of HAN on the Twitter dataset (unpaired t-test, $p = 0.096$). OOT: Out of Time (24hrs).}
	\begin{tabular}{l | cc | cc | cc | cc}
			& \multicolumn{2}{c|}{Amazon} & \multicolumn{2}{c|}{Twitter} & \multicolumn{2}{c|}{YouTube} & \multicolumn{2}{c}{Tissue-PPI} \\
		\hline
		\rule{0pt}{2ex}				& ROC-AUC & F1 	& ROC-AUC & F1 & ROC-AUC & F1 & ROC-AUC & F1 \\
		\hline
		\rule{0pt}{2ex} node2vec	& 94.47 		 & 87.88 		  & 72.58		   & 71.94 			& 71.21 		 & 65.36		  & 51.30		   & 64.04 \\
						DeepWalk	& 94.20 		 & 87.38 		  & 76.88 		   & 72.42 			& 71.11 		 & 65.52		  & 58.48		   & 67.16 \\
		\hline
		\rule{0pt}{2ex} MNE			& 90.28			 & 83.25 		  & OOT			   & OOT			& 82.30 		 & 75.03 		  & OOT 		   & OOT \\
						R-graphSAGE	& 94.88 		 & 89.39 		  & 74.31	  	   & 70.77			& 87.02 		 & 79.93 		  & 66.61 		   & 61.59 \\
						R-GCN		& 94.96			 & 90.08		  & 92.75		   & 85.85			& 80.21			 & 73.36 		  & 84.19		   & 75.98 \\
						GATNE		& 96.25 		 & 91.36 		  & 92.94 		   & 86.20 			& 84.47 		 & 76.83 		  & 79.83 		   & 71.78 \\
						HAN			& 95.28			 & 90.43		  & \textbf{94.81} & \textbf{88.44} & 80.43 		 & 73.43 		  & 93.05		   & 85.98 \\
						RAHMeN		& \textbf{96.78} & \textbf{92.39} & \textbf{94.58} & \textbf{88.31}	& \textbf{88.64} & \textbf{80.58} & \textbf{94.88} & \textbf{87.99} \\ 
	\end{tabular}
\label{tab:results}
\end{table*}

\subsection{Baseline Comparisons} \label{baselines}

We evaluate RAHMeN against two categories of baselines: \emph{homogeneous network embedding models} and \emph{multiplex heterogeneous network embedding models}. The embedding dimension for all models is set to 200; the complete hyperparameter settings for each model are listed in the Appendix. We utilize two well-established graph embedding methods originally designed for homogeneous networks: node2vec~\cite{node2vec} and DeepWalk~\cite{deepwalk}. To adapt these methods to AHMeNs, we learn node embeddings for each separate relation layer in the multiplex networks. We compare RAHMeN against five state-of-the-art AHMeN embedding frameworks: MNE~\cite{mne}, R-GCN~\cite{rgcn}, R-graphSAGE, GATNE~\cite{gatne}, and HAN~\cite{han}. We implement an R-graphSAGE framework following the R-GCN design principles and use graphSAGE in place of GCN. R-GCN, R-graphSAGE, and HAN aggregate messages from relation-specific graph convolutional operations---GCN~\cite{kipf_gcn}, graphSAGE~\cite{graphsage}, and GAT~\cite{gat}, respectively. By contrast, MNE and GATNE both learn a common base node embedding, which is augmented with an edge-specific embedding that captures the information contained in various edge types. MNE applies a fixed weight to each individual edge-type message, while GATNE implicitly learns aggregation weights when combining the individual edge-type messages. For the inductive experiment, we compare RAHMeN against R-graphSAGE, GATNE, and HAN since these models support both inductive learning and attributed networks.

In contrast to the heterogeneous embedding frameworks listed above, RAHMeN learns an inductive transformation function that leverages both node attributes and the multi-relational structure of the network. At each $k$-level in the model, we learn relation-specific convolutional operations that transform and combine a node's self-attributes and its local context in the network. We then employ self-attention to share information across relation types in the network, learning a set of latent representations of a node that capture the multi-faceted semantic contexts of a node in AHMeNs. Specific implementation details may be found in the appendix. We evaluate the performance of RAHMeN by comparing against the aforementioned baseline models on link prediction experiments in both transductive and inductive settings.

\subsection{Transductive Experiment Results} \label{transductive}

For our experiments in a transductive setting, we use a classic link prediction task, which is common in both academic and industrial contexts. In our experimental setup, we mask a set of edges and non-edges from the original network and train the models on the remaining network. We utilize the train/val/test splits provided by \cite{gatne} for the Amazon and YouTube datasets. For the Twitter dataset, we follow the same procedure and create validation and test sets that consist of 5\% and 10\% randomly selected positive edges, respectively, with an equivalent number of randomly selected negative edges for each relation type. For the Tissue-PPI dataset, we use 5-fold cross validation with 20\% of the edges in the network held-out and split into validation and test sets. We report the area under the ROC curve (ROC-AUC) and the F1 score of all models on the link prediction task. To avoid the thresholding effect, we assume the number of hidden edges in the test set is given~\cite{threshold,gatne}. We report the mean performance over five trials for each dataset; both metrics are uniformly averaged over all relation types in the datasets.

\begin{figure*}[!ht]
	\centering
	\includegraphics[width=0.9\textwidth]{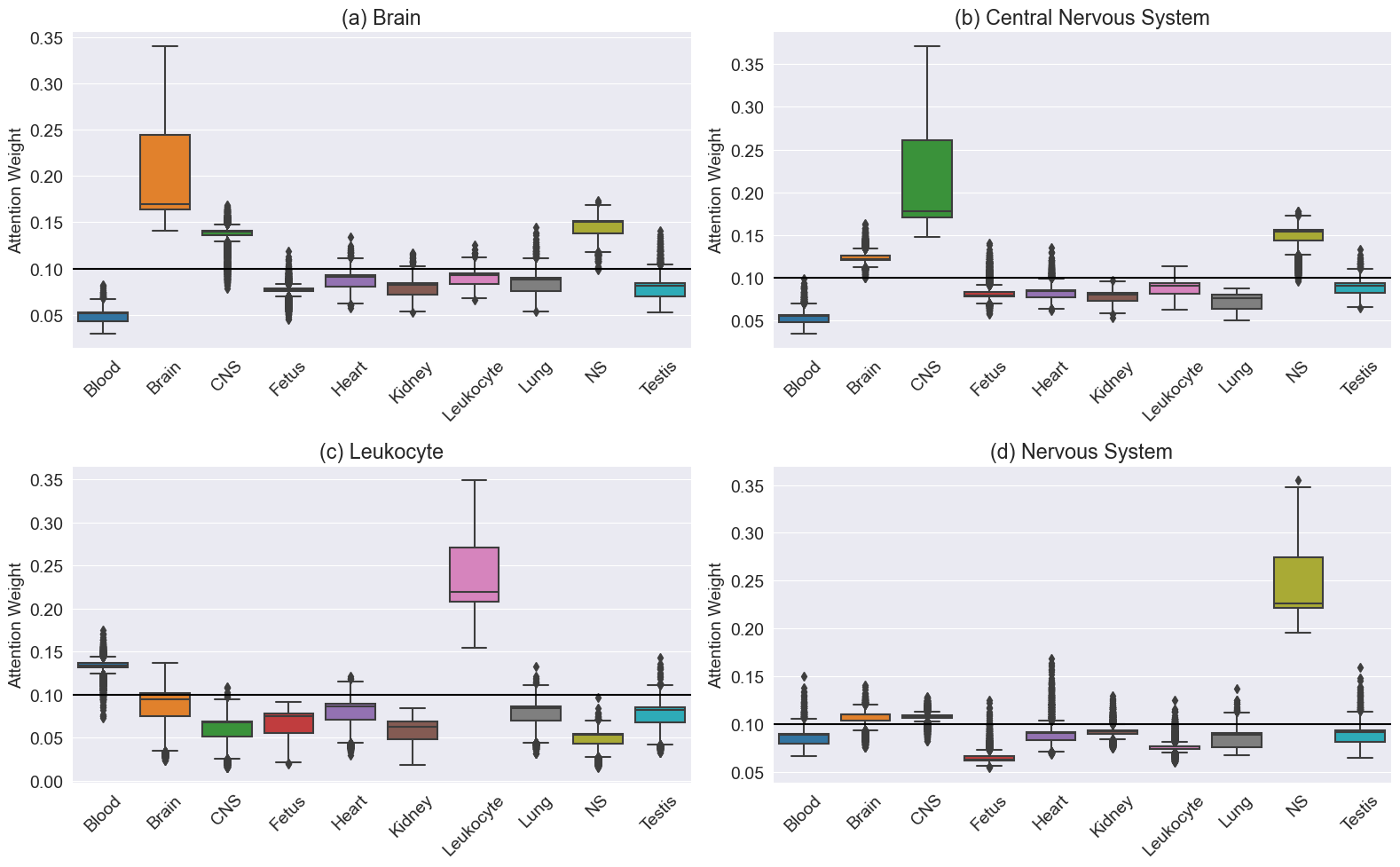}
	\caption{Distributions of tissue self-attention values for each protein for the (a) Brain, (b) CNS, (c) Leukocyte, and (d) NS layers in the Tissue-PPI dataset shows the significance of self-attention mechanism of the RAHMeN framework in identifying biologically related tissues.}
	\label{fig:attn}
  \end{figure*}

The experimental results are presented in Table~\ref{tab:results}. The results clearly illustrate the strong performance of RAHMeN across all four of the AHMeN datasets. RAHMeN outperforms all baseline models on the Amazon, YouTube, and Tissue PPI datasets (unpaired t-test, $p < 0.05$) and achieves comparable results to HAN on the Twitter dataset (unpaired t-test, $p = 0.096$). MNE did not complete training on the Twitter and Tissue-PPI datasets due to the out of time issue (24hrs). We note that the baseline models, excluding RAHMeN, demonstrate variable performance across the experimental datasets. Each of the baseline frameworks, aside from GATNE, generate a single embedding to represent a node in an AHMeN, and GATNE relies largely on its base self-node embedding, as the edge-type information is highly compressed into a low-dimension space. This fact limits the expressiveness of other heterogeneous embedding frameworks and requires that these models compromise on the extent to which they emphasize the self-node or emphasize the multi-relational local context of the node. As such, depending on the particular characteristics of the network. RAHMeN, by contrast, demonstrates consistently strong performance across all of the experimental datasets. In the RAHMeN framework, we learn a transformation function that computes multi-embeddings for a target node which integrates a multi-faceted view of the self-node's attributes and the multi-relational structure of the network. This provides RAHMeN with a greater expressive capacity than other heterogeneous graph embedding models, producing latent representations of nodes that better capture the diverse interactions between entities in AHMeNs. Overall, our experiments demonstrate that RAHMeN is able to integrate multi-relational local graph structures in the network and to share information across layers producing information-rich embeddings that capture the complex structure of attributed heterogeneous multiplex networks.

\subsection{Inductive Experiment Results} \label{inductive}

In addition to our experiments in a transductive setting, we conduct an evaluation of RAHMeN in an inductive context on the Tissue-PPI dataset. For this experiment, we mask 15\% of nodes in the graph to consider as test nodes unseen during model training. From the reduced graph, we mask an additional 20\% randomly sampled positive edges with an equivalent number of randomly sampled negative edges as a validation set for hyperparameter tuning and early stopping. For evaluation, we create a test graph from the training graph by adding 50\% of the edges incident on the hidden nodes which provide the local structure of the hidden nodes during neighbor sampling. Our link prediction task is to predict the remaining 50\% of edges from the unseen nodes, with an equivalent number of randomly sampled negative edges incident on the hidden nodes. Table~\ref{tab:inductive_results} illustrates our inductive experiment results, which demonstrate that RAHMeN maintains its performance advantage over all other heterogeneous graph embedding frameworks in an inductive context (unpaired t-test, $p < 0.05$). 

\subsection{Self-Attention Explainability} \label{attention}

The relational self-attention is critical for the enhanced expressiveness of RAHMeN as it facilitates information sharing across the varied contexts present in AHMeNs. In addition to improving the  performance of RAHMeN embeddings on downstream learning tasks, we hypothesize that our relational self-attention discovers interpretable connections in the network, which provides a level of explainability to RAHMeN that is not present in other heterogeneous graph embedding frameworks. To further examine our attention mechanism, we analyze the learned attention weights for each relation in the Tissue-PPI dataset where each relation layer in this dataset corresponds to a particular human tissue type. For our experiments we extract the graph consisting of the ten largest tissue layers in the network. The relational layers are: blood, brain, central nervous system (CNS), fetus, heart, kidney, leukocyte, lung, nervous system (NS), and testis. Figure~\ref{fig:attn} illustrates the distribution of attention values for every protein node in the network with respect to each tissue type in our dataset, and as Figure~\ref{fig:attn} demonstrates, we can see that RAHMeN's self-attention mechanism identifies the importance of biologically-related tissues in predicting protein interactions. Figure~\ref{fig:attn}a shows the attention distributions for the brain layer, where we note that both the CNS and nervous system layers are attended to by the model when learning latent representations of proteins in the brain. Similarly, Figure~\ref{fig:attn}b shows that attention distributions for the CNS layer, where both the brain and NS layers are important. By contrast, Figure~\ref{fig:attn}d shows that the more general nervous system layer does not attend as highly over the brain and CNS layers compared to the other tissue layers. Lastly, Figure~\ref{fig:attn}c shows the attention values for the leukocyte layer, where we see that white blood cell proteins attend to other proteins found in the blood. In all, our analysis of RAHMeN's learned attention weights in the Tissue-PPI model illustrates that our relational self-attention produces interpretable and biologically consistent importance weights across the tissue types present in the protein-protein interaction network.

\begin{table}
	\centering
	\caption{Performance (ROC-AUC \% and F1 \%) of embedding frameworks on the link prediction task in an inductive context. RAHMeN outperforms all baseline models (unpaired t-test, $p < 0.05$)}
	\begin{tabular}{ l | c|c }
		\rule{0pt}{2ex} & \multicolumn{2}{c}{Tissue-PPI} \\
					\hline
					& ROC-AUC & F1 \\
					\hline
		\rule{0pt}{2ex}R-graphSAGE & 71.16 & 65.46 \\
		R-GCN 		& 78.08 & 70.98 \\
		GATNE 		& 61.36 & 57.64 \\
		HAN 		& 87.53 & 57.66 \\
		RAHMeN 		& \textbf{87.93} & \textbf{79.94} \\
	\end{tabular}
	\label{tab:inductive_results}
\end{table}

\subsection{Graph Motif Features} \label{motifs}

\begin{figure}
	\centering
	\includegraphics[width=7cm]{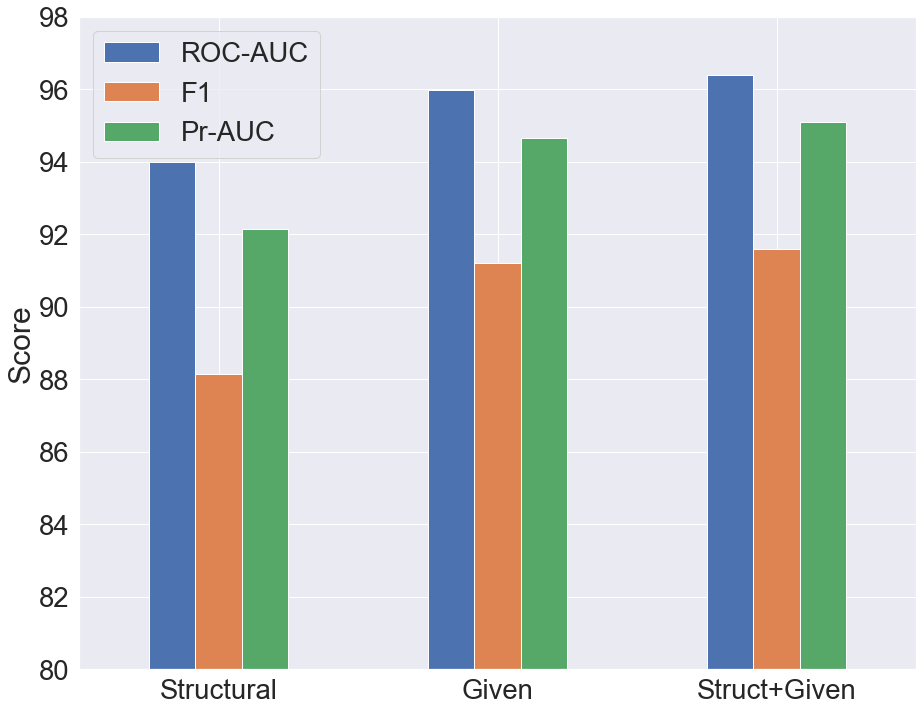}
	\caption{Performance of RAHMeN on the Amazon dataset with structural (motif) features, given node features, and a combination of given and structural features.}
	\label{fig:motif_perf}
\end{figure}

Meaningful initial feature representations are essential for generating high-quality embeddings with inductive graph embedding models. Such models learn an inductive transformation function that embeds a node in the latent feature space based on the characteristics of the self-node and feature distributions of a node's local neighborhood. In many cases, initial node features must be generated through transductive embedding frameworks or other models. An attractive alternative is to generate initial feature representations from graph structures both as initial node representations or as a method of feature augmentation. In this work, we generate structural feature representations by counting the participation of each node in various network motifs. Motifs are small graph substructures that appear with regular frequency in a graph~\cite{motifDef}. In our experiment, we consider connected motifs containing two, three, or four nodes. We limit the counted motifs to four nodes because the number of motif types grows exponentially as you increase the number of participating nodes. Diagrams of the three and four node motifs are shown in Figure~\ref{fig:motifs}. To obtain motif counts, we use a modified version of the graphlet counting algorithm presented by~\cite{graphletCounting}. Our modifications allow us to count motif participation with respect to each individual node in the graph as opposed to producing the total count for each motif type in the graph. We account for network heterogeneity by counting motif participation on each relation layer independently. Our results with the motif-based node features are shown in Figure~\ref{fig:motif_perf} and indicate that structural representations can augment node features or serve as a viable alternative to learning initial node embeddings using DeepWalk~\cite{deepwalk} or other transductive embedding frameworks, as was the case for the YouTube and Twitter datasets~\cite{gatne}.

\subsection{Ablation Study \& Parameter Sensitivity} \label{params}

\begin{figure}
	\centering
	\includegraphics[width=8cm]{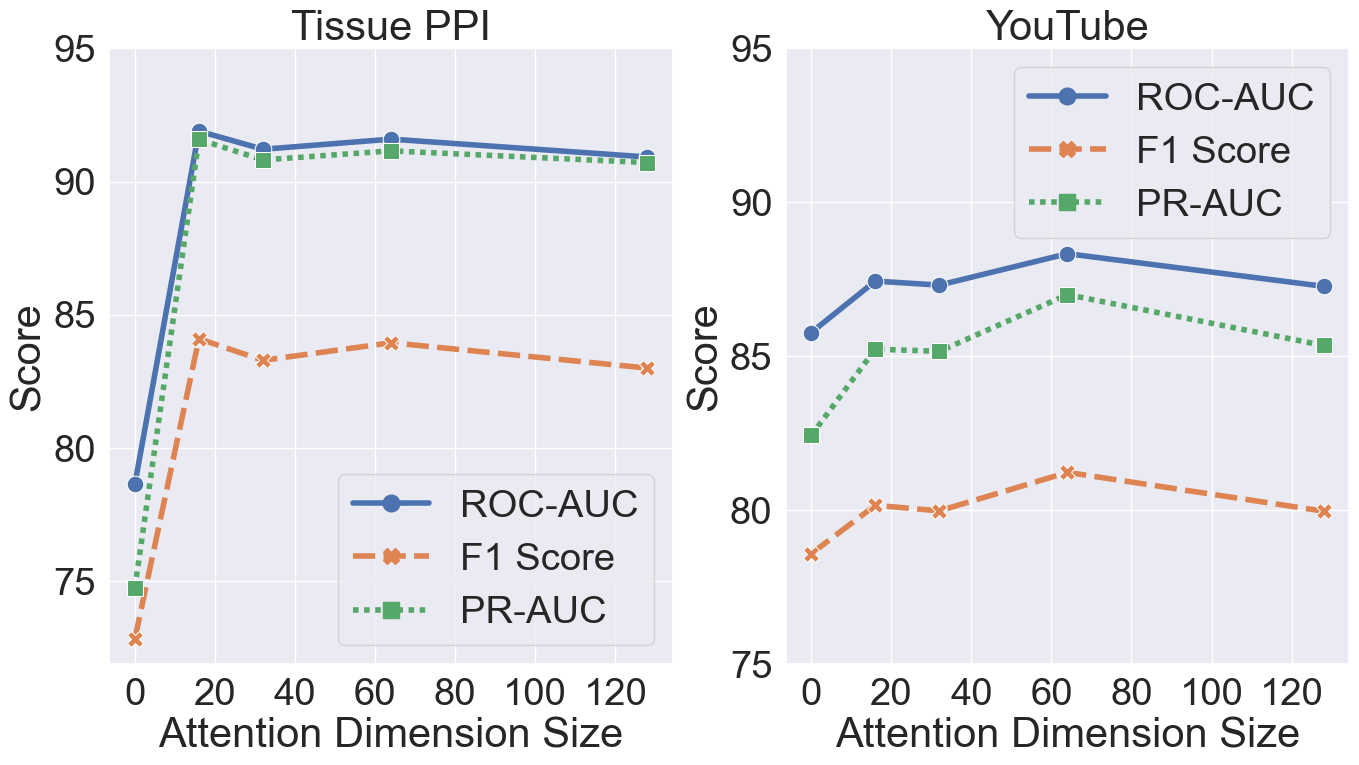}
	\caption{Ablation study and dimension analysis of RAHMeN's relational self-attention. For the Tissue-PPI dataset, information sharing across relations is highly important for model performance.}
	\label{fig:ablat}
  \end{figure}

To demonstrate the necessity of RAHMeN's attention mechanism, we conduct an ablation test where RAHMeN's relational self-attention is removed from the model. In this variant, the relation-specific graph convolution operations transform and aggregate information about a node's self-attributes and its local graph neighborhood. This information is then propagated through the model along each relation axis without any sharing of contextual information from the other relations present in the network. Figure~\ref{fig:ablat} illustrates the performance of RAHMeN on the YouTube and Tissue-PPI with no self-attention. As can be seen, removal of RAHMeN's relational self-attention reduces the overall performance of the model on both datasets; though, the extent of the performance decrease varies between the two datasets. In the Tissue-PPI dataset, where network relations correspond to biological tissues, removing the information sharing from the self-attention leads to a significant drop in performance. This aligns with biological knowledge that many proteins are present in multiple tissues and participate in functionally similar roles across different tissues---a fact reinforced by our analysis of RAHMeN's attention presented in Section ~\ref{attention}. In the YouTube dataset, the drop in performance is more muted, suggesting that cross-relational information sharing is less significant for users in the YouTube social network.

\begin{figure*}
	\centering
	\includegraphics[width=\textwidth]{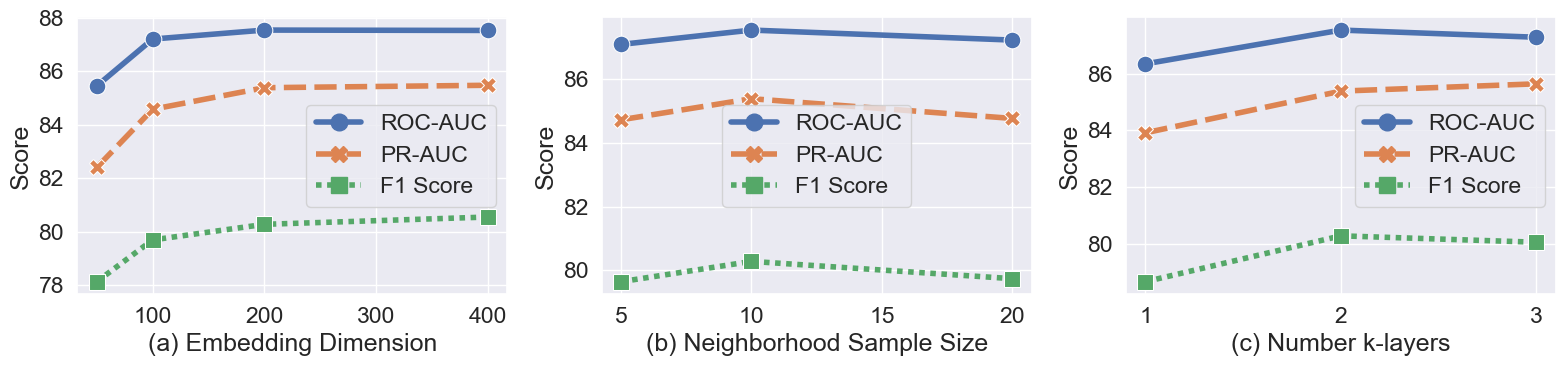}
	\caption{\small{The sensitivity of RAHMeN when varying the (a) size of the embedding dimension, (b) modifying the size of the neighborhood sample at each $k$-level, and (c) altering the number of $k$-level neighbor aggregations. We observe that RAHMeN is stable within a large range of embedding sizes and insensitive to neighborhood sample size; though performance drops when the embedding dimension is too small. A single level of neighborhood aggregation results in lower performance compared to two levels of aggregation. Three levels of aggregation does not result in a significant increase in performance, and does not offset the resulting increase in computational costs.}}
	\label{fig:param_sensitivity}
\end{figure*}

We also investigate the sensitivity of RAHMeN to various hyperparameters, including embedding dimension $d$, neighborhood sample size, number of $k$-levels, and self-attention dimension. Figure~\ref{fig:param_sensitivity}a illustrates the performance of RAHMeN when changing the size of the embedding dimension from which we can conclude that the performance of RAHMeN is stable within a large range of embedding sizes; though performance drops when the embedding dimension is too small. Figure~\ref{fig:param_sensitivity}b illustrates RAHMeN's performance when modifying the size of the neighborhood sampled per node at each level in the model. The performance of RAHMeN when altering the number of $k$-level convolutional layers is depicted in Figure~\ref{fig:param_sensitivity}c, and we can see that a single level of neighborhood aggregation results in lower performance compared to two levels of aggregation. Three levels of aggregation does not result in a significant increase in performance, and does not offset the resulting increase in computational costs. 

\section{Conclusion}
In this work, we introduce RAHMeN: a framework for relation-aware embeddings for attributed heterogeneous multiplex networks. Our framework applies the graph convolutional approaches to learn a node's spatial embedding with respect to each canonical relation in the network. RAHMeN allows embeddings to be generated for unseen nodes by training a set of relation-specific graph convolutional operators that learn a set of relation-specific characterizations for each node. By incorporating relational semantic self-attention, which uncovers the importance of each relation to one another and facilitates sharing information across all relations in the network, RAHMeN produces a set of information-rich multi-embeddings for each node in the network that capture the diverse nature of nodes in AHMeNs. 

RAHMeN outperforms state-of-the-art benchmarks on the link prediction in both transductive and inductive contexts in datasets in both social and biological settings. We also demonstrate the utility and interpretability of RAHMeN's self-attention over the relations present in a network, and we show the potential of incorporating low-level subgraph features as node attributes. A number of potential future directions are possible for RAHMeN including incorporating node level attention in neighborhood aggregation, motif-based attention for local structure-aware neighborhood sampling, and extensions to higher-order structures such as hypergraphs.

%\section*{Acknowledgment}

%The preferred spelling of the word ``acknowledgment'' in America is without 
%an ``e'' after the ``g''. Avoid the stilted expression ``one of us (R. B. 
%G.) thanks $\ldots$''. Instead, try ``R. B. G. thanks$\ldots$''. Put sponsor 
%acknowledgments in the unnumbered footnote on the first page.
\vspace{1cm}
\bibliographystyle{IEEEtran}
\bibliography{bibliography}

% Appendix
\newpage
\appendix
\section{Appendix}

In the appendix, we describe the implementation details of our proposed model along with details of the experiments conducted. We provide detailed descriptions of datasets and the parameter configurations for all methods utilized herein.

\subsection{Implementation Details}

The experiments were conducted using a Windows PC with an Intel(R) Core(TM) i7-7600 CPU @ 2.80GHz, 32GB RAM and an NVIDIA 2060 Super-8GB. Our models were implemented using PyTorch 1.8.1\footnote{https://pytorch.org/} and DGL 0.6.1\footnote{https://www.dgl.ai/} in Python 3.8. Our experimental code may be separated into two components: random walk generation and model training/evaluation. The random walk component of our model uses DGL's heterogeneous network random walk functionality with reference to GATNE's PyTorch implementation\footnote{https://github.com/THUDM/GATNE} and a reference implementation of GATNE-T using DGL\footnote{github.com/dmlc/dgl/tree/master/examples/pytorch/GATNE-T}. These references along with references to graphSAGE\footnote{https://github.com/williamleif/GraphSAGE} and a reference implementation of graphSAGE in DGL\footnote{https://github.com/dmlc/dgl/tree/master/examples/pytorch/graphsage} were utilized to develop our model implementation and to develop the training procedure. We utilize the RelGraphConv implementation of R-GCN in DGL\footnote{https://docs.dgl.ai/api/python/nn.pytorch.html\#relgraphconv}, and the DGL implementation of HAN along with the author's original repository\footnote{https://github.com/Jhy1993/HAN}. Our evaluation of model performance uses functions from scikit-learn\footnote{https://scikit-learn.org/stable/}, including \textit{roc\_auc\_score}, \textit{f1\_score}, \textit{precision\_recall\_curve}, and \textit{auc}. Model parameters are optimized using stochastic gradient descent and updated using the Adam optimizer~\cite{adam}. For the Amazon and YouTube results, we report the results as presented by \cite{gatne} and utilize the same model parameters for our experiments on the Twitter and Tissue-PPI datasets.

\paragraph{Parameter Configuration.} We set the hidden and embedding dimension $d$ for all experiments to 200. The number of random walks for each node is 20, and the length of each walk is 10. The sliding window size for node contexts is set to 5. We use 5 negative samples for each training sample. The number of training epochs is capped at 50, and the model will stop early if the validation ROC-AUC does not improve for 3 consecutive epochs. The dimension of the attention used in the model is set to 20. We utilize the default Adam optimizer with learning rate set to 0.001. For the Twitter dataset, the number of k-levels was set to 1. For the remaining experiments, K was set to 2 levels of aggregation.

\subsection{Baseline Comparisons}
For all baseline models, the embedding size is set to 200. For random-walk methods we use 20 random walks of length 10 with sliding window of size 5. The number of skip-gram iterations is set to 100. We use the author's code repository for node2vec, DeepWalk, MNE, and GATNE, and the DGL implementations of R-GCN, R-graphSAGE, and HAN. Training and evaluation for all graph neural network models follows the same random-walk based procedure described in Section \ref{modelopt}.

\begin{itemize}
	\item \textbf{node2vec}\cite{node2vec}. We utilize the code from the author's GitHub repository.\footnote{https://github.com/aditya-grover/node2vec} Parameter $p$ is set to 2 and parameter $q$ is set to 0.5. 
	\item \textbf{DeepWalk}\cite{deepwalk}. We utilize the code from the author's GitHub repository. \footnote{https://github.com/phanein/deepwalk}
	\item \textbf{R-graphSAGE}\cite{graphsage}. We implement a graphSAGE model in DGL using the DGL SAGEConv layer.
	\item \textbf{R-GCN}\cite{rgcn}. We implement an R-GCN model in DGL using the DGL RelGraphConv Layer.
	\item \textbf{MNE}\cite{mne}. We utilize the code from the author's GitHub repository. \footnote{https://github.com/HKUST-KnowComp/MNE}
	\item \textbf{GATNE}\cite{gatne}. We utilize the code from the author's GitHub repository. \footnote{https://github.com/THUDM/GATNE}
\end{itemize}

\subsection{Datasets}

We utilize four publicly available datasets for our experiments. Table \ref{tab:origdatasets} describes the statistics of the original datasets. We utilize the processed datasets for Amazon and YouTube from \cite{gatne} and utilize their train/val/test splits. For the Amazon and YouTube experiments, results for node2vec, DeepWalk, MNE, and GATNE were reproduced from \cite{gatne}. Experiments with R-GCN, R-graphSAGE, HAN, and RAHMeN were repeated for five trials. For the Twitter dataset, similarly five trials were conducted using our train/val/test split for each model variant. For the Tissue-PPI dataset, the transductive experiments were conducted with 5-fold cross validation. Because of the necessity of training a separate model for each layer of the network, experiments were conducted using a single cross validation split for DeepWalk and node2vec. MNE did not finish training within 24 hours for the Twitter and Tissue-PPI datasets. For the inductive experiment, five trials were repeated for R-GCN, R-graphSAGE, GATNE, HAN, and RAHMeN.

\begin{table}
	\centering
	\caption{Original Datasets}
	\begin{tabular}{l | rr | l}
		\textbf{Dataset} & \# nodes & \# edges & \# relations \\
		\hline
		Amazon		& 312,320	& 7,500,100		& 4		\\
		Twitter		& 456,626	& 15,367,315	& 4		\\
		YouTube		& 15,088	& 13,628,895  	& 5		\\
		Tissue-PPI	& 4,510		& 3,666,563 	& 107	\\
	\end{tabular}
	\label{tab:origdatasets}
\end{table}

%\vspace{12pt}
%\color{red}
%IEEE conference templates contain guidance text for composing and formatting conference papers. Please ensure that all template text is removed from your conference paper prior to submission to the conference. Failure to remove the template text from your paper may result in your paper not being published.

\end{document}